# Automated Landfill Detection Using Deep Learning: A Comparative Study of Lightweight and Custom Architectures with the AerialWaste Dataset

**Nowshin Sharmily[1], Rusab Sarmun[1], and Muhammad E. H. Chowdhury[2,\*], Senior Member, IEEE, Mir Hamidul Hussain[3], Saad Bin Abul Kashem[4], Molla E Majid[5], Amith Khandakar[2], Senior Member, IEEE,**

[1]Department of Electrical and Electronic Engineering, University of Dhaka, Dhaka, Bangladesh. Email: nowshinsharmilyeee@gmail.com; rusabsarmun@gmail.com.
[2]Department of Electrical Engineering, College of Engineering, Qatar University, Doha, Qatar. Email: mchowdhury@qu.edu.qa; amitk@qu.edu.qa
[3]Department of Computer Science and Engineering, School of Data and Sciences, Brac University, Dhaka, Bangladesh. Email: mirhamidul635@gmail.com
[4]Department of Computing Science, AFG College with the University of Aberdeen, Doha, Qatar. Email: saad.kashem@afg-aberdeen.edu.qa
[5]Computer Applications Department, Academic Bridge Program, Qatar Foundation, Doha, Qatar. Email: mmajid@qf.org.qa (MEM)

* Corresponding author: Muhammad E. H. Chowdhury (mchowdhury@qu.edu.qa)

The open access publication of this article is supported by Qatar National Library (QNL).

**ABSTRACT** Illegal landfills are posing as a hazardous threat to people all over the world. Due to the arduous nature of manually identifying the location of landfill, many landfills go unnoticed by authorities and later cause dangerous harm to people and environment. Deep learning can play a significant role in identifying these landfills while saving valuable time, manpower and resources. Despite being a burning concern, good quality publicly released datasets for illegal landfill detection are hard to find due to security concerns. However, AerialWaste Dataset is a large collection of 10434 images of Lombardy region of Italy. The images are of varying qualities, collected from three different sources: AGEA Orthophotos, WorldView-3, and Google Earth. The dataset contains professionally curated, diverse and high-quality images which makes it particularly suitable for scalable and impactful research. As we trained several models to compare results, we found complex and heavy models to be prone to overfitting and memorizing training data instead of learning patterns. Therefore, we chose lightweight simpler models which could leverage general features from the dataset. In this study, Mobilenetv2, Googlenet, Densenet, MobileVit and other lightweight deep learning models were used to train and validate the dataset as they achieved significant success with less overfitting. As we saw substantial improvement in the performance using some of these models, we combined the best performing models and came up with an ensemble model. With the help of ensemble and fusion technique, binary classification could be performed on this dataset with 92.33% accuracy, 92.67% precision, 92.33% sensitivity, 92.41% F1 score and 92.71% specificity.

## I. INTRODUCTION

The exponential increase in global waste production and illegal landfills have led to significant environmental and health concerns. These unchecked and illegal landfills do not have adequate safeguards which causes substantial harm to the environment by contaminating the soil and water which can eventually pose serious health hazard to the inhabitants of the neighboring area including both humans and wildlife. Landfills contaminate groundwater with hazardous leachates containing toxic organic compounds and mutagenic substances, posing acute and chronic health risks like cancer [1]. Properties near illegal dump sites often experience a decline in value. Studies have found that proximity to hazardous waste sites can lead to property price reductions ranging from 2% to 8%. This depreciation affects homeowners' equity and reduces property tax revenues for local governments, further impacting community resources [2]. The development of landfills near residential areas is often linked to issues such as methane (CH₄) gas accumulation, groundwater contamination, and property damage [3].

Illegal dumping sites release significant amounts of methane, a potent greenhouse gas, as organic waste decomposes anaerobically. Methane is 84 times more effective at absorbing the sun's heat than carbon dioxide, making it a substantial contributor to global warming. To find illegal dumping sites is key to protecting the environment and can help slow down the effects of global warming and climate change [4-6]. Therefore, Identifying and monitoring landfills, especially illegal dumping sites, is crucial for mitigating environmental hazards, enforcing regulations, and improving waste management policies [7, 8]. Traditional methods for landfill detection rely on manual on-site inspections and photo interpretation, which are time-consuming, resource-intensive, and prone to human error [9]. Besides, it is quite a challenge to manually identify small illegal dumpsites by using traditional methods. As a result, these sites often go unnoticed until they grow significantly, typically being reported by residents or detected by authorities only after reaching a substantial volume [10]. Criminal organizations illegally incinerate waste to destroy evidence of hazardous materials, releasing toxic fumes like dioxins that pose severe public health risks. This uncontrolled burning exacerbates environmental contamination, increasing the toxicity of air, soil, and groundwater [11]. Municipal microdumps, often compounded by illegal manufacturers, can exacerbate the surroundings of urban areas[12].

Satellite imaging technology has emerged as a powerful tool for Earth observation over the past decade. Numerous studies have explored its use in locating dumpsites, employing both manual and semi-automated methods to enhance detection accuracy [13]. With the rise of artificial intelligence and computer vision, powered by better computing capabilities and a growing amount of annotated data, building accurate image classification models has become more feasible than ever. However, applying these models to waste detection remains challenging due to the need for a large set of training images. To create reliable machine learning models for landfill classification, it's crucial to collect a substantial number of landfill locations along with their aerial images. The advent of deep learning and remote sensing technologies has enabled automated and scalable approaches to waste site detection using aerial and satellite imagery [14-16].

This study explores the effectiveness of deep learning models in detecting landfills using the AerialWaste dataset, which consists of high-resolution aerial and satellite images annotated with waste site information [17]. Various lightweight and custom deep learning architectures have been evaluated, including MobileNetV2 and Vision Transformer (ViT) to classify images as containing waste (positive) or not (negative)[18-20]. For binary classification ensemble methods were utilized, as they are supposed to give better results in classification compared to single classifiers[21-23]. The study also introduces ensemble models and explores the impact of different optimization strategies on classification performance.

In this work, our contributions are four-fold:

1) Benchmarking multiple deep learning models for landfill detection using the AerialWaste dataset.
2) Investigating the impact of different optimizers on classification performance.
3) Proposing ensemble models that combine CNN-based and transformer-based architectures to enhance detection accuracy.
4) Evaluating computational efficiency and real-world applicability of lightweight models for large-scale environmental monitoring.

This paper addresses the gap in automated landfill detection methods by focusing on improving the accuracy and efficiency of detecting illegal landfills in diverse aerial and satellite images. Though in recent years, there has been influx of aerial images, automated detection systems have yet to reap the benefit of it. Despite existing models, challenges such as dataset imbalance, variations in image quality, and the difficulty of distinguishing waste from surrounding environments remain. The research aims to enhance model generalization and reduce overfitting by leveraging lightweight architectures and ensemble techniques for better detection in complex and resource-constrained settings.
.

The remainder of this paper is structured as follows: Section 2 reviews related work on automated landfill detection and deep learning applications in waste monitoring. Section 3 overviews the methodology, including data preprocessing, model training, and evaluation metrics. Section 4 presents the experimental results, ablation studies, and performance comparisons between several deep learning models. Finally, Section 5 discusses key findings, limitations, and future research directions.

## II. Literature Review

Remote sensing technology and Earth observation satellites with medium and high-resolution multispectral sensors have advanced significantly in recent decades. These modern instruments can capture data across both visible and invisible wavelengths, offering a more comprehensive view of the observed area and enhancing our understanding of the environment [24]. Satellite imagery and GIS technologies provide a reliable means of mapping, tracking, and analyzing waste sites which help governments to detect and regulate illegal waste dumping effectively. Aerial photographs are widely used to identify disposal sites, and remote sensing techniques allow for monitoring structural changes in hazardous waste sites over time. Meanwhile, the conventional approach of waste monitoring and regulation can be financially burdensome. For example, the U.S. Environmental Protection Agency (USEPA) reported that in 2000, businesses incurred nearly $32 billion in costs to comply with hazardous waste regulations, highlighting the significant financial burden of conventional waste monitoring approaches [**25**]

Interpretation of remote sensor data and aerial photos can yield a variety of hydrological and soils data for identification of potential landfill sites [**26**]. Remote sensing offers a cost-effective and efficient way to identify and monitor landfills and solid waste disposal sites. Simultaneously, satellite imagery enables us to have large-scale coverage, reduces over reliance on expensive infrastructure, prioritizes human investigations on high-risk locations. It also facilitates repeated observations of the same area to track site evolution over time which is also a time-consuming job. Researchers have developed specialized techniques that analyze different signs of landfill presence, such as visible waste, unique spectral patterns, vegetation stress, and surface temperature. These methods help identify both large landfills and smaller dumping sites, monitor existing waste locations, and evaluate suitable areas for new landfill development [**17**].

In recent years, neural networks are being used as the basis of deep learning in the field of remote sensing. Since deep learning algorithms have gained massive success lately in many image analysis tasks including land use and land cover (LULC) classification, scene classification, and object detection, the remote-sensing community has shifted its attention to deep learning [**27-30**].

Fraternali et al., in their paper, trained and validated the AerialWaste dataset using a ResNet50 backbone augmented with Feature Pyramid Network (FPN) links during training[**31**]. Though the aerial images obtained from various sources (Google Earth, WorldView, AGEA) exhibit variability in resolution and quality, it also hampers consistent performance across different datasets. Again, there is imbalance in data, as there are far too many negative examples available compared to the positive ones. While the dataset's technical validation is carried out using a CNN model (ResNet50 with FPN), a more thorough comparison with other architectures or an exploration of potential areas of improvement could provide a clearer understanding of why this specific architecture was chosen and its performance in comparison to others.

Table 1 represents relevant literature in this field with methodology, dataset and key findings:

TABLE I
REVIEW OF RELEVANT LITERATURE

| Study | Publication Year | Dataset | Methodology |
|---|---|---|---|
| Fraternali et al. [31] | 2023 | AerialWaste Dataset | Dataset Development & Benchmarking |
| Fraternali et al.[9] | 2024 | Multiple datasets from literature | Literature Review |
| Molina et al. [32] | 2024 | AerialWaste Dataset | Image Super-Resolution & Deep Learning |
| Padubidri et al.[15] | 2022 | High-resolution aerial images from Houston, USA | CNN-based classifier (MobileNetV2) |
| Rajkumar et al. [16] | 2022 | WorldView-3, WorldView-2, GeoEye-1 satellite images | U-Net, FCN, ResNet34 segmentation |
| Torres & Fraternali [17] | 2021 | ~3,000 aerial images from Italy (20 cm resolution per pixel) | ResNet50 + Feature Pyramid Network (FPN) |
| Sun et al. [10] | 2023 | 2,500 labeled dumpsites (~4800 sq. km, 28 cities worldwide) | BCA-Net (deep learning classifier) |
| Niu et al. [33] | 2022 | VHR remote sensing imagery from Google Earth (Total 3680 image patches) | Integrating a multi-scale dilated CNN and a Swin-Transformer |

## III. Methodology

The classification of waste sites from aerial and satellite images using the AerialWaste dataset was carried out in three stages- Data Processing, Model Training and Evaluation and Analysis. The whole process is summarized in Figure 1.

### A. Data Processing

The process leveraged the AerialWaste dataset, with training and testing datasets curated from JSON files containing metadata and annotations. A total of 11,703 images were extracted, with separate folders created for training and testing purposes. A total of 7,278 images were allocated to the training folder. From this folder, 20% of the images (1,818) were set aside to create a validation folder, ensuring balanced evaluation during model training. The remaining 2,607 images were assigned to the testing folder. All images were resized to a standard resolution of 256×256 pixels to ensure uniformity across the dataset.

The training dataset initially had an imbalanced number of positive and negative images. To ensure that the positive and negative folders of the training set have equal number of images, the positive set was augmented rendering both of the folders with 4852 images.

After balancing, the positive images were further augmented using various transformations such as random rotation, horizontal and vertical flipping, color jitter, and random cropping. This augmentation process aimed to improve model accuracy and reduce overfitting, not to balance the classes. By rotating images, the model learned to detect waste regardless of its spatial alignment in the image, enhancing robustness against real-world variations in angle and orientation. Flipping is used to simulate mirrored waste scenarios, as waste can appear in any direction within the image. The color jitter technique helps simulate different lighting conditions and variations in image quality that might arise from different aerial image sources. It also ensures that the model doesn't become too sensitive to specific lighting or color distributions. The augmented positive images contributed to better generalization, helping the model perform better on unseen data.

Apart from these changes, we made some basic changes in ground truth. When we went through the annotated images thoroughly, some images were wrongly classified in the training set. So, we had to sort it again to shift them in the right class. For example, in Figure 2, we can clearly see rubbles and containers, but primarily these images were classified as negative. So, after reviewing the images, we put them in the positive class. In Figure 3, there are images of clean fields and forest, but these pictures were classified as positive ones. So, we relocated them in negative class upon review**.**

While the augmentation methods can help simulate some of these variations, the inherent differences in image quality might still introduce bias**.**

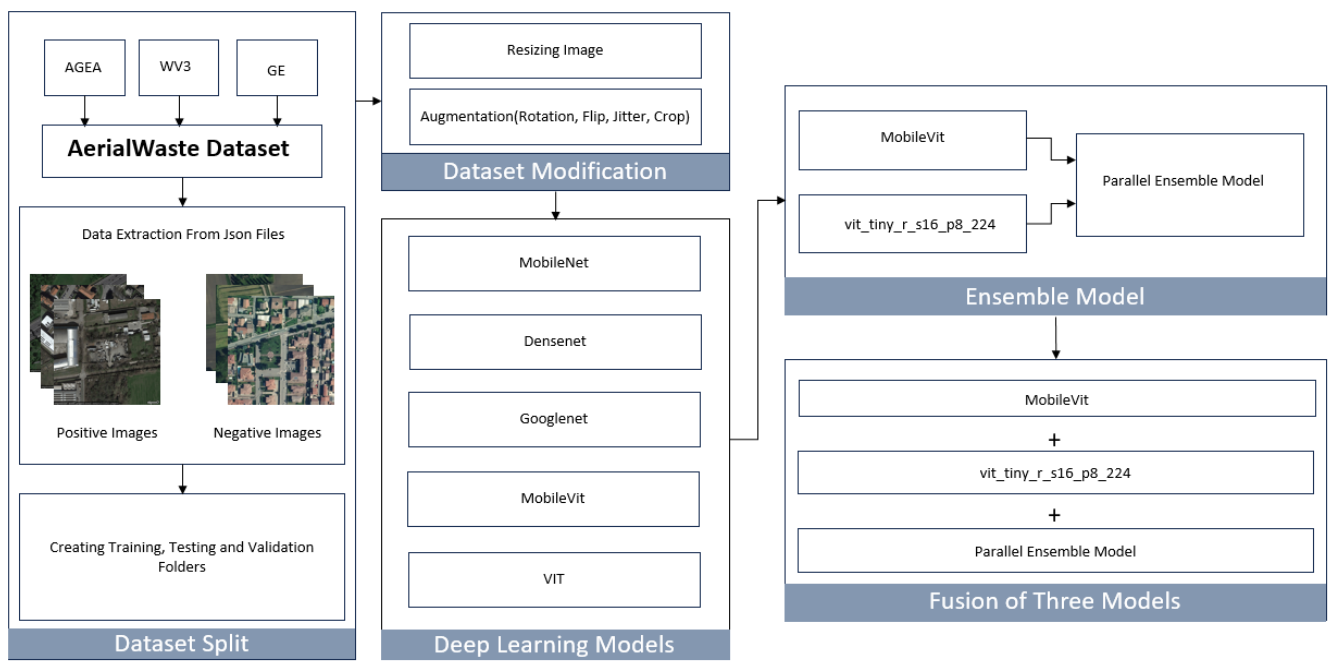

FIGURE 1. Data Processing and Ensemble Model

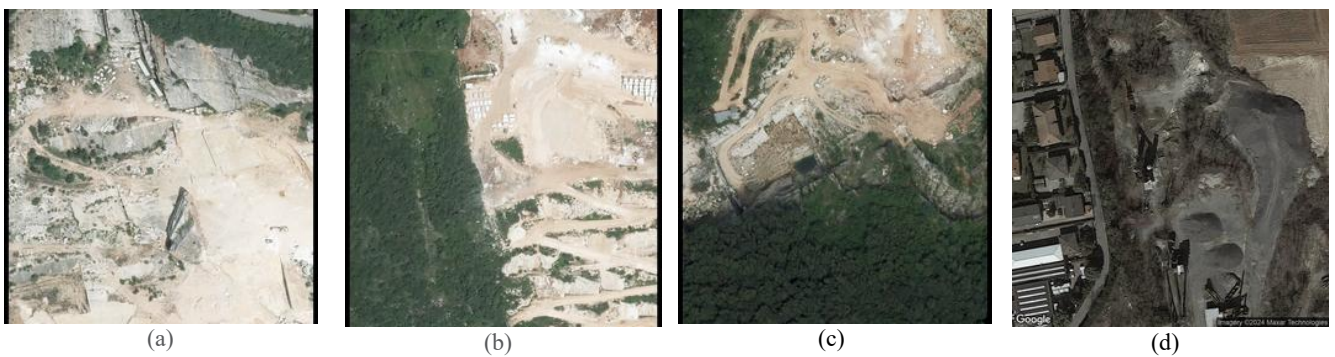

FIGURE 2. Images that were classified as negative, but should be classified as positive.

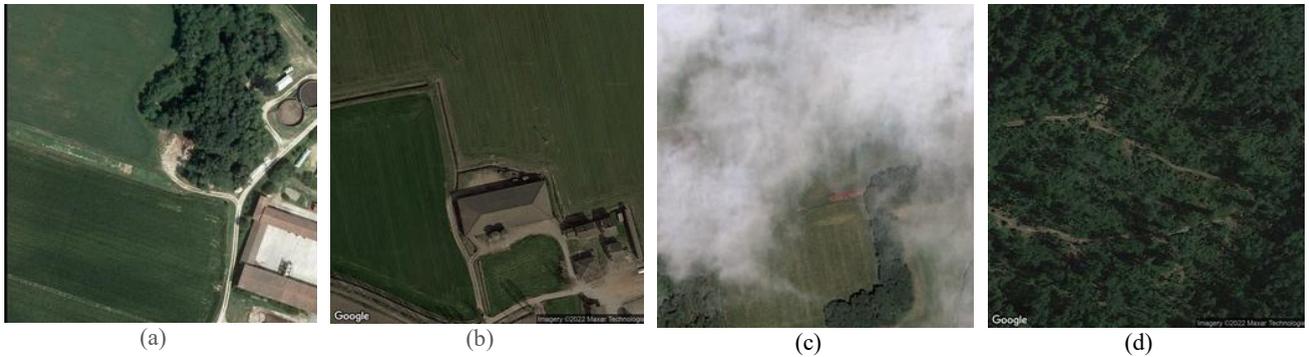

(a)        (b)        (c)        (d)

**FIGURE 3.** Images that were classified as positive, but should be classified as negative.

### B. Model Training

Several lightweight and custom deep learning architectures were explored for classifying images as containing waste (positive) or not (negative). The following models were employed:

#### 1) MOBILENETV2 (VARIANTS)

MobileNetV2 is a convolutional neural network architecture designed for efficient image classification [18]. The MobileNetV2 model was initialized with weights pre-trained on ImageNet, a large and diverse dataset containing millions of labeled images across various categories. This pre-training allows the model to learn fundamental visual features that are common across many image recognition tasks. Recent studies have demonstrated that applying transfer learning and model pruning techniques to MobileNetV2 can further enhance its performance on large image datasets. For instance, customizing the model's architecture and employing pruning strategies have led to improved accuracy and reduced model size, facilitating efficient processing of bulk images [34]. MobileNetV2_050 (width multiplier 0.5) and MobileNetV2_100 (width multiplier 1.0), with pretrained weights and frozen initial layers were used to train separately and in parallel ensemble models.

MobileNetV2-0.5 is a lightweight deep learning model designed for efficient object detection and classification, particularly in resource-constrained environments. It builds upon the MobileNetV2 architecture, which introduces inverted residual blocks with linear bottlenecks to improve computational efficiency while maintaining high accuracy [18]. The "0.5" variant refers to a reduced width multiplier, making it even more compact and suitable for mobile and edge applications. MobileNetV2 employs depthwise separable convolutions, significantly reducing the number of parameters and FLOPs while maintaining strong feature extraction capabilities [35]. Due to its optimized structure, MobileNetV2-0.5 achieves an effective trade-off between accuracy and computational efficiency, making it an ideal choice for real-time AI applications on mobile and embedded systems. The lightweight nature, quick processing and low memory usage makes it a perfect choice for landfill detection.

#### 2) MOBILEVIT-XS

MobileViT-XS is a compact yet powerful deep learning architecture that integrates convolutional neural networks (CNNs) with transformer-based feature extraction to enhance both local and global feature representation. Designed for lightweight vision tasks, it retains the efficiency of traditional convolutional models while leveraging the strengths of self-attention mechanisms. Unlike conventional CNNs, which rely solely on local receptive fields, MobileViT-XS applies transformers to process image patches globally, allowing it to capture long-range dependencies more effectively. This hybrid design makes it particularly well-suited for applications requiring computational efficiency, such as real-time object detection, mobile vision tasks, and, as explored in this study, automated landfill detection from aerial and satellite imagery [35]. Like MobileNetV2, MobileViT-XS is also pretrained on ImageNet. Figure 4 explains the architecture of MobileVit XS with a block diagram.

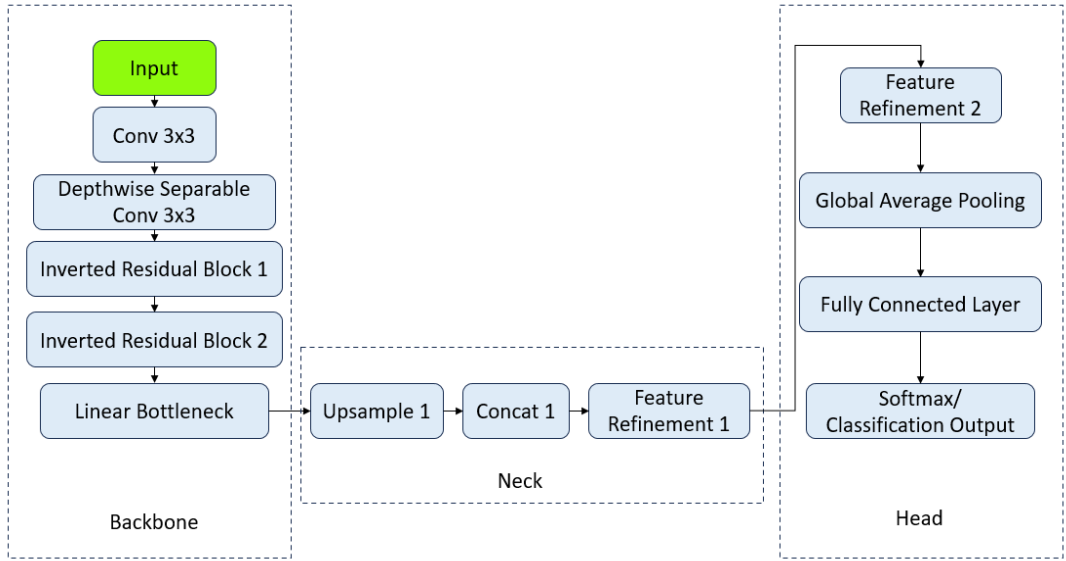

**FIGURE 4.** MobileVit XS Architecture

One of the defining characteristics of MobileViT-XS is its depthwise separable convolutions, which significantly reduce computational complexity while preserving spatial feature integrity. Traditional convolutional layers involve costly operations that scale poorly in resource-constrained environments. By replacing them with depthwise separable convolutions, MobileViT-XS minimizes the number of parameters and floating-point operations (FLOPs), making it highly efficient for mobile and embedded systems. Additionally, the model integrates transformer-based processing by dividing feature maps into non-overlapping patches and feeding them into a multi-head self-attention (MHSA) module, a mechanism first introduced in the Vision Transformer (ViT) architecture [19]. This approach enables MobileViT-XS to excel in long-range dependency learning, improving its ability to recognize complex spatial structures in images.

Another critical component of the MobileViT-XS architecture is its feature fusion mechanism, which merges outputs from transformer layers with convolutional features. This fusion allows the model to maintain high spatial resolution while benefiting from the global contextual understanding provided by self-attention. After processing through the transformer encoder, the refined features are reintroduced into the CNN backbone, which enhances both fine-grained local information and broad contextual understanding. This dual-processing capability enables MobileViT-XS to achieve better generalization across diverse datasets, making it robust in applications where both local texture patterns and global scene understanding are crucial [36].

In the context of this study, MobileViT-XS played a crucial role in an ensemble model alongside ViT-Tiny, significantly improving landfill detection accuracy. While traditional CNN models like DenseNet121 and GoogLeNet demonstrated strong feature extraction capabilities, they lacked the ability to capture global dependencies effectively. By integrating MobileViT-XS with ViT-Tiny, the ensemble model leveraged both local and global feature representations, leading to superior classification performance. The ability of MobileViT-XS to process high-resolution aerial images efficiently, while maintaining lightweight computational requirements, further underscored its suitability for real-world environmental monitoring applications [20].

Overall, MobileViT-XS stands out as an optimal solution for tasks requiring both computational efficiency and high classification accuracy. By combining the advantages of CNNs and transformers, it effectively balances speed, accuracy, and resource utilization, making it an ideal candidate for large-scale automated waste detection initiatives. As advancements in transformer-based architectures continue to evolve, further refinements to MobileViT-XS, such as adaptive feature fusion and multi-scale input processing, could enhance its capabilities even further.

### 3) VISION TRANSFORMER (VIT):

Lightweight variants such as ViT Tiny (e.g., vit_tiny_r_s16_p8_224) is a compact version of the Vision Transformer (ViT) architecture, designed to offer high efficiency in image recognition tasks with reduced computational cost. Unlike traditional convolutional neural networks (CNNs), ViT-Tiny applies a self-attention mechanism to process images as sequences of patches, effectively capturing long-range dependencies between features. The "S16" in its name denotes a 16×16 patch size, while "P8" refers to an 8×8 patch embedding stride. It ensures a balance between computational efficiency and feature representation [35]. Instead of relying on conventional convolutional feature extractors, ViT-Tiny models global relationships between pixels, enabling it to achieve strong generalization performance on diverse image datasets. It is particularly advantageous for applications where transformer-based architectures are preferred but computational resources are limited. Same as the previous two models, ViT models are pretrained on ImageNet, which enables it to leverage transfer learning while recognizing patterns.

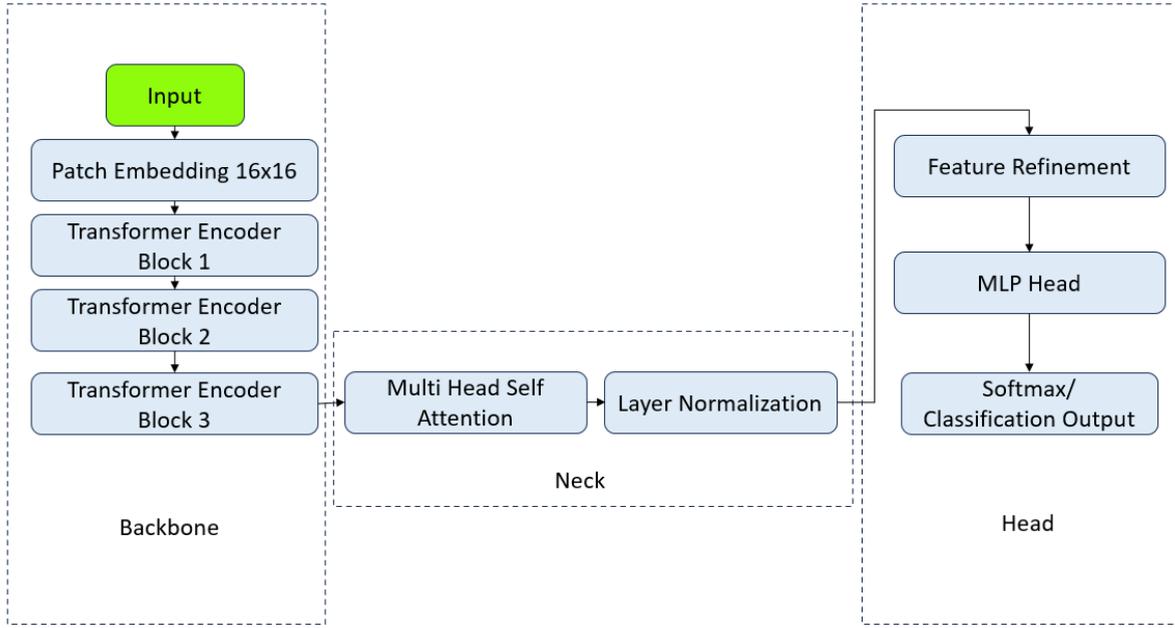

**FIGURE 5.** ViT-Tiny-R-S16-P8-22 Architecture

The architecture of ViT-Tiny includes three major components: the Backbone (Feature Extraction), the Neck (Feature Aggregation), and the Head (Classification Output). The backbone begins with a patch embedding layer, where the input image is tokenized into a sequence of patches. After that multiple Transformer Encoder Blocks are organized, each block has Multi-Head Self-Attention (MHSA), Feedforward MLP layers, and Layer Normalization (LN). The Neck section is used for enhancing feature representation by refining embeddings through additional self-attention layers and normalization to stabilize training. Lastly, the Head consists of a fully connected MLP layer and a Softmax classifier, which determines probabilities for each class [35]. This architecture is explained via three blocks naming backbone, neck and head in Figure 5.

### C. Matrices for Model Evaluation

In assessing the performance of the classification task, this study incorporates a diverse set of metrics. These metrics include Accuracy, Precision, Sensitivity (Recall), F1-score, and Specificity, all of which are derived from the confusion matrix

$$Accuracy = \frac{TP + TN}{TP + TN + FP + FN} \quad (1)$$

$$Specificity = \frac{TN}{TN + FP} \quad (2)$$

$$Sensitivity = \frac{TP}{TP + FN} \quad (3)$$

$$Precision = \frac{TP}{TP + FP} \quad (4)$$

$$F1\ Score = 2 \times \frac{Precision \times Sensitivity}{Precision + Sensitivity} \quad (5)$$

where, TP, TN, FP and FN denote true positive, true negative, false positive, and false negative, respectively.

The blend of accuracy, specificity, sensitivity, precision and F1-score provide a well-rounded assessment. Since both kinds of misclassification are concerning in this case, the F1-score offers the most balanced and holistic measure of the model's performance.

### D. Experimental Setup for Model Training for Model Evaluation

This multi-model and ensemble approach provided a robust foundation for landfill detection, balancing computational efficiency with classification performance. This section outlines the experimental setup used for training and evaluating the Aerialwaste Parallel Ensemble Model, ensuring consistency across all runs. For training and evaluating our models, we used Google Colab with GPU acceleration, which made processing high-resolution aerial images much more efficient. The GPU significantly cut down on training time, allowing for quicker iterations and better model tuning. This cloud-based platform offered a scalable environment for managing large datasets like AerialWaste, ensuring that deep learning models ran smoothly [37].

The model was trained using a batch size of 16 and a single-fold cross-validation strategy (num_folds=1). To standardize input preprocessing, the dataset was normalized using per-channel mean and standard deviation values: input_mean = [0.3201, 0.3334, 0.2832] and input_std = [0.2004, 0.1818, 0.1764], respectively. The Negative Log-Likelihood Loss (NLLLoss) function was employed for training, and input images were resized to 224×224 pixels before feeding them into the model. The training process included 10 inference steps (N_steps=10). The training parameters for the binary classification model are detailed in Table II. Key settings included a batch size of 64, a learning rate of 1e-4, and a maximum of 100 epochs.

TABLE II
TRAINING PARAMETERS FOR BINARY CLASSIFICATION

| Training parameters | Parameter value |
|---|---|
| Batch size | 64 |
| Learning rate | 1e-4 |
| Number of folds | 1 |
| Input Channel | 3 |
| Max epochs | 100 |
| Epoch patience | 20 |

### E. Loss Function

In this study, Negative Log-Likelihood Loss (NLLLoss) was used as the primary loss function for training the classification models. Given that some of the architectures output log-probabilities (instead of raw logits), NLLLoss was chosen for its numerical stability and compatibility with log-softmax outputs [38]. The use of NLLLoss ensured that the model penalized incorrect predictions based on confidence levels, effectively guiding optimization [39].

Given a dataset with input samples x and corresponding ground truth labels y, the model outputs a probability distribution P(y|x). The NLLLoss function is defined as:

$$L = -\sum_{i=1}^{N} \log P(y_i|x_i) \qquad (6)$$

where $P(y_i|x_i)$ is the predicted probability of the correct class for sample $i$, $N$ is the total number of samples and the log function ensures that lower confidence predictions result in higher penalty values.

### F. Other Models and Ablation Study

A comparative study was conducted on various architectures, including DenseNet121, SqueezeNet1_0, and GoogLeNet, to evaluate their performance in binary classification tasks. To further enhance accuracy, ensemble techniques were introduced, such as Parallel Ensemble Models, which combined Mobilevit_xs and ViT Tiny outputs for improved predictions. Additionally, models were trained using different optimization strategies, including AdamW, RAdam, Ranger, and SGD with Warm Restarts, to fine-tune performance. Transfer learning was leveraged by initializing models with pretrained weights, followed by fine-tuning specific layers to adapt to the task. To optimize computational efficiency, mixed precision training was employed, reducing overhead without sacrificing performance. Furthermore, in some cases, specific layers (e.g., the first 10 layers) were frozen to accelerate training and retain essential pretrained features while enabling the model to adapt effectively to the dataset.

### G. Evaluation and Analysis

The trained models were evaluated on a held-out test set using metrics such as accuracy, precision, recall, F1 score. The performance of different models was compared to determine the most effective architecture for landfill detection.

## IV. Result

This section presents the evaluation of the proposed models for landfill detection, highlighting their performance on key metrics such as accuracy, precision, recall, and F1-score. Results for individual models, the ensemble approach, and feature extraction with classical machine learning are detailed below.

Although we did augmentation to balance the dataset, the inherent imbalance affected the training process, which caused misclassified images. Also, the deeper architectures were prone to overfitting, as the dataset was quite large. So, lightweight models, benefiting significantly from transfer learning could lessen misclassified instances.

### A. Training with Different Models

The comparative evaluation of different deep learning models for landfill detection highlights the effectiveness of lightweight architectures over deeper models. Among the tested models, MobileNetV2_050 with 10 frozen layers achieved the highest accuracy of 90.68%, along with strong precision (90.71%), sensitivity (90.68%), and F1-score (90.69%), demonstrating its ability to generalize well while maintaining efficiency.

The standard MobileNetV2_050 model without frozen layers also performed well, achieving 89.8% accuracy, 89.96% precision, and an F1-score of 89.86%. This suggests that even without additional optimization, MobileNetV2 effectively balances performance and computational efficiency. SqueezeNet1_0, another lightweight model, achieved 89.34% accuracy with a competitive 89.91% precision and 89.47% F1-score, making it a strong alternative.

Among the deeper models, DenseNet121 reached 89.14% accuracy, while GoogLeNet had a slightly lower accuracy of 88.34%. Despite their strong feature extraction capabilities, these models did not outperform the more compact architectures, possibly due to overfitting or the additional computational complexity required for training.

Overall, the results emphasize the advantage of lighter models like MobileNetV2 and SqueezeNet, which provided higher accuracy and better generalization while being computationally efficient. Their streamlined architectures and depthwise separable convolutions allowed them to achieve strong performance without excessive model size or training time, making them well-suited for real-world applications where efficiency is crucial. Table 3 explains the performance of different deep learning models with the prepared dataset in terms of accuracy, precision, sensitivity, F1 score and specificity.

TABLE III
PERFORMANCE OF DIFFERENT CNNS IN BINARY CLASSIFICATION

| Model | Accuracy | Precision | Sensitivity | F1 score | Specificity |
|---|---|---|---|---|---|
| Densenet_121 | 89.14 | 89.31 | 89.14 | 89.20 | 87.60 |
| mobilenetv2_050 | 89.8 | 89.96 | 89.79 | 89.86 | 88.45 |
| Googlenet | 88.34 | 88.25 | 88.34 | 88.14 | 82.46 |
| Squeezenet1_0 | 89.34 | 89.91 | 89.34 | 89.47 | 89.66 |
| mobilenetv2_050(10 frozen layers) | 90.68 | 90.71 | 90.68 | 90.69 | 88.66 |

### B. Analysis of Different Models

The ParallelEnsembleModel (mobilevit_xs_vit_tiny_r_s16_p8_224) is an ensemble model designed to combine the strengths of MobileViT-XS and ViT Tiny (ViT-Tiny-R-S16-P8-224) for classification. MobileViT-XS is a hybrid model that integrates CNN and transformer-based processing, making it efficient for both spatial feature extraction and global context modeling, while ViT Tiny specializes in capturing long-range dependencies using self-attention. Both models are initialized with pretrained weights and fine-tuned for the classification task. During the forward pass, an input image is independently processed by both models, and their outputs are concatenated rather than averaged, allowing the network to retain richer feature representations. The concatenated output is then passed through a fully connected layer (fc1), which learns to optimally fuse the representations from both networks. This approach provides greater flexibility compared to simple averaging, enabling the model to determine the most informative features from each backbone. The ensemble design enhances classification accuracy by leveraging the complementary strengths of MobileViT's efficient hybrid structure and ViT Tiny's powerful attention mechanisms.

These results indicate that the ensemble model not only outperformed individual architectures but also provided better robustness in handling class imbalances and challenging cases in the AerialWaste dataset. In Table 4, the performance matrices clearly show the parallel ensemble model outperforming both mobilevit_xs and vit_tiny_r_s16_p8_224 in case of accuracy, precision, sensitivity, F1 score and specificity.

The ensemble model effectively combines MobileViT-XS and ViT Tiny (ViT-Tiny-R-S16-P8-224), leveraging their complementary strengths in spatial feature extraction and global context modeling. This ensemble approach achieved an accuracy of 91.56% and an F1-score of 91.59%, outperforming both individual models—MobileViT-XS (91.25% accuracy, 91.37% F1-score) and ViT Tiny (90.95% accuracy, 90.98% F1-score). The model benefits from MobileViT-XS's hybrid CNN-transformer structure, which efficiently balances local and long-range dependencies, while ViT Tiny enhances the model's ability to capture global relationships. The ensemble's precision (91.65%) and sensitivity (91.56%) further indicate strong generalization and robustness, though its specificity (90.25%) is slightly lower than that of MobileViT-XS (91.88%). The improved performance suggests that concatenating feature representations and refining them through a fully connected layer allows the model to learn a richer, more informative feature fusion compared to simple averaging. Future enhancements could explore adaptive weighting mechanisms for fusion, multi-scale feature integration, or transformer-based self-attention refinement to further improve classification performance.

TABLE IV
PERFORMANCE COMPARISON OF INDIVIDUAL CNNS WITH THE PARALLEL ENSEMBLE MODEL

| Model | Accuracy | Precision | Sensitivity | F1 score | specificity |
|---|---|---|---|---|---|
| ParallelEnsembleModelmobile vit_xs_vit_tiny_r_s16_p8_224 | 91.56 | 91.65 | 91.56 | 91.59 | 90.25 |
| mobilevit_xs | 91.25 | 91.74 | 91.26 | 91.37 | 91.88 |
| vit_tiny_r_s16_p8_224 | 90.95 | 91.04 | 90.95 | 90.98 | 89.49 |

### C. ROC Plot of The Parallel Ensemble Model

A Receiver Operating Characteristic (ROC) curve is a graphical representation used to evaluate the performance of a binary classification model by plotting the True Positive Rate (TPR) against the False Positive Rate (FPR) across various threshold settings. ROC analysis is one of multiple tools used for understanding CNN models and their test results, as a goal of explainable artificial intelligence [39]. The Area Under the Curve (AUC) quantifies the overall ability of the model to discriminate between positive and negative classes, with a value of 1 indicating perfect discrimination and 0.5 suggesting performance no better than random chance [40]. ROC curves are particularly useful in assessing the trade-offs between sensitivity and specificity [41].

Figure 6 shows the True Positive Rate (TPR) vs. False Positive Rate (FPR) across different classification thresholds. Here, the dashed diagonal line represents a random classifier (AUC = 0.5), and any curve above it indicates a model performing better than random guessing. As the classifier in question shows high AUC (0.97), it has outperformed random classifier significantly. The pink dashed line represents Micro-Average AUC and the blue dotted line represents Macro-Average AUC. The minimal gap between micro- and macro-average AUC values suggests that the model maintains consistency across different class distributions.

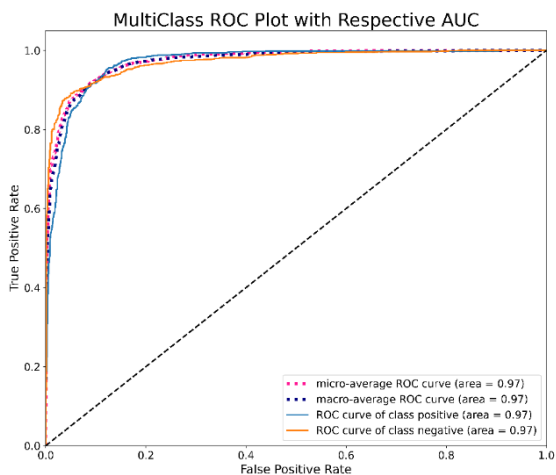

**FIGURE 6.** ROC plot of the ParallelEnsembleModel

An AUC score of 0.97 indicates that the model performs exceptionally well in distinguishing between classes. The ROC curves are close to the top-left corner (TPR = 1, FPR = 0), which suggests that the model achieves high sensitivity (recall) with minimal false positives. The high TPR and low FPR indicate that the model achieves a good balance between sensitivity and specificity, making it a viable solution for real-world waste detection using remote sensing.

### D. Fusion of Three Models to Improve Performance

To enhance classification performance, we used an ensemble-based approach by combining predictions from three different models: ParallelEnsembleModel_mobilevit_xs_vit_tiny_r_s16_p8_224, MobileViT-XS, and ViT-Tiny-R-S16-P8-224. Each model generated class probabilities for the positive and negative classes, which were stored in separate files. The predictions from these models were then aggregated by matching images based on their filenames and computing the average probabilities for both classes. After obtaining the final predictions, we evaluated model performance using standard classification metrics, including accuracy, precision, sensitivity, F1-score, and specificity. The results demonstrated that the ensemble approach improved overall classification accuracy compared to individual models. Table V presents the detailed performance metrics of this fusion approach. All prediction outputs and computed metrics were saved for further analysis.

### E. Ablation Study: Comparing Different Optimizers and Their Impacts Model Performance

To understand the impact of applying different optimization strategies on landfill detection, we conducted an ablation study using various optimizers with our ParallelEnsembleModel. The objective was to assess how optimizer choice affects the parameters like as accuracy, precision, recall, and F1-score. The comparison of the performance is shown in Table 6.

The ablation study on different optimizers for landfill detection revealed significant performance variations in terms of accuracy, precision, sensitivity, F1-score, and specificity. Among the tested optimizers, AdamW achieved the highest accuracy (91.14%) and F1-score (91.26%), demonstrating its robustness in optimizing model weights efficiently. Ranger, a blend of RAdam and Lookahead optimizers, showed competitive performance with an accuracy of 91.06% and an F1-score of 91.12%, making it a strong alternative

.

### F. A Comparison with Existing Approaches

In recent years, there have been significant work in CNN guided illegal waste site detection. Table VII shows few studies that are closely comparable to our approach. Fraternali et al. and M. Molina et al. worked on the same dataset, but M. Molina's use of super-resolution for image enhancement gave it a significant edge over Fraternali et al.'s approach. None of them used lightweight models for ensemble which approach gave our work a new direction.

TABLE V
PERFORMANCE MATRICES OF DIFFERENT CLASSES

| CLASS | Accuracy | Precision | Sensitivity | F1 score | specificity |
|---|---|---|---|---|---|
| Positive | 92.33 | 85.23 | 93.09 | 88.99 | 91.95 |
| Negative | 92.33 | 96.38 | 91.95 | 94.11 | 93.09 |
| Weighted Average | 92.33 | 92.67 | 92.33 | 92.41 | 92.71 |

TABLE VI
PERFORMANCE MATRICES OF DIFFERENT CLASSES

| OPTIMIZER | Accuracy | Precision | Sensitivity | F1 score | specificity |
|---|---|---|---|---|---|
| AdamW | 91.14 | 91.71 | 91.14 | 91.26 | 92.05 |
| Ranger | 91.06 | 91.25 | 91.06 | 91.12 | 90.23 |
| RAdam | 90.07 | 90.07 | 90.07 | 89.88 | 84.42 |
| Rprop | 89.72 | 89.75 | 89.72 | 89.73 | 87.36 |
| SGD with Warm Restarts | 89.64 | 89.81 | 89.64 | 89.70 | 88.26 |

TABLE VII
PERFORMANCE MATRICES OF DIFFERENT CLASSES

| REFERENCE | Dataset | Approach | Performance |
|---|---|---|---|
| Fraternali et al. | AerialWaste dataset (3,478 positive and 6,956 negative images) | ResNet50 backbone augmented with Feature Pyramid Network (FPN) links | 87.99% average precision and 80.70% F1 score |
| M. Molina et al. | AerialWaste dataset | Super-Resolution enhancement on downscaled images | Accuracy 0.9044, Precision 0.8416, Recall 0.8793, F1 score 0.8600 |
| A. Rajkumar et al | Private Dataset (High resolution multi-spectral satellite images fromWorldView-3, WorldView-2 and GeoEye-1 | Pretrained CNNs on SpaceNet | Accuracy 0.83, Precision 0.645, Recall 0.550, Specificity 0.919, |
| S. Lin et al.[32] | The CWLD dataset | The DeepLabV3 + network | Accuracy 0.96, Precision .88, Recall 90.70, F1 score 89.01 |

## V. Conclusion

The automated detection of landfills from aerial and satellite imagery is a crucial step toward enhancing environmental monitoring and waste management efforts. This study highlights the effectiveness of deep learning models in automated landfill detection, with a particular focus on ensemble modeling, which significantly improved classification performance over individual architectures. The ParallelEnsembleModel, combining MobileViT-XS and ViT-Tiny, outperformed all standalone models, achieving the highest accuracy (91.56%), precision (91.65%), and F1-score (91.59%). By leveraging the strengths of CNN-based feature extraction and transformer-based global context modeling, the ensemble model demonstrated superior generalization and robustness in detecting landfill sites across diverse aerial and satellite imagery.

Compared to individual models like MobileNetV2, SqueezeNet, DenseNet121, and GoogLeNet, the ensemble approach provided better feature representation by merging outputs from different architectures. The complementary nature of convolutional and self-attention mechanisms allowed for richer spatial and contextual feature learning, reducing misclassifications and improving sensitivity to subtle patterns in the images. Additionally, the fusion of multiple optimizers, including AdamW and Ranger, further refined performance by balancing stability and convergence.

Despite the success of the ensemble model, challenges remain in improving adaptability to varying environmental conditions and image resolutions. Future work should explore adaptive weighting mechanisms for feature fusion, multi-scale learning, and self-supervised training to enhance classification accuracy. Incorporating multi-modal datasets and expanding the training data across different geographical regions that contains diverse weather conditions can further strengthen the generalizability of the model.

In conclusion, this study demonstrates that ensemble modeling is a powerful approach for landfill detection, surpassing individual deep learning architectures in both accuracy and efficiency. By refining ensemble techniques and integrating more advanced fusion strategies, future developments can make automated landfill monitoring even more reliable and scalable for real-world environmental applications.

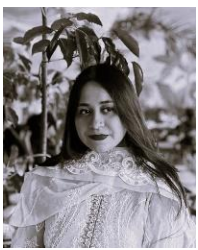
**Nowshin Sharmily** received her B.Sc. and M.Sc. degrees from University of Dhaka in Electrical and Electronic Engineering. Her research interests include Machine learning, Computer Vision, and Biomedical Engineering.

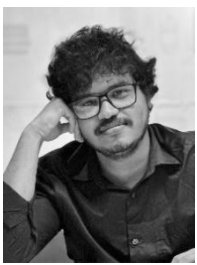
**Rusab Sarmun** received the Bachelor of Science degree in Electrical and Electronic Engineering from the University of Dhaka. He is currently working as a Research Assistant in the Machine Learning Group at Qatar University, where his research primarily focuses on the application of deep learning in computer vision, with a strong emphasis on the biomedical domain. His research interests include medical image analysis, deep learning model development, and computer vision for disease detection and diagnosis. He has contributed in several high-impact Q1 publications focusing on the detection and diagnosis of diseases such as chronic kidney disease, pulmonary embolism, diabetic foot ulcers, Ischemic stroke, Lumbar spine Segmentation, and prostate cancer grading.

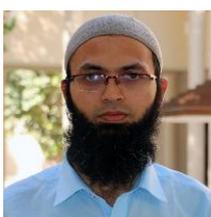
**Muhammad E. H. Chowdhury** received his Ph.D. from the University of Nottingham, U.K., in 2014. He subsequently worked as a Postdoctoral Research Fellow at the Sir Peter Mansfield Imaging Centre, University of Nottingham. Currently, he serves as an Assistant Professor and Program Coordinator in the Department of Electrical Engineering at Qatar University. Dr. Chowdhury is a prolific researcher with several patents to his name, two edited books, and over 200 peer-reviewed journal articles, 30+ conference papers, and multiple book chapters. His research interests span biomedical instrumentation, signal processing, wearable sensors, medical image analysis, machine learning, computer vision, embedded system design, and simultaneous EEG/fMRI. He is actively involved in leading several research projects funded by the Qatar Research, Development, and Innovation Council (QRDI) and internal grants from Qatar University, along with academic collaborations with HBKU and HMC. A Senior Member of IEEE, Dr. Chowdhury also serves as an Associate Editor for Computers and Electrical Engineering and IEEE Access, as well as a Topic Editor and Review Editor for Frontiers in Neuroscience. Dr. Chowdhury has been recognized with several prestigious awards, including the COVID-19 Dataset Award, AHS Award from HMC, and the National AI Competition award for his contributions to the fight against COVID-19. His team earned a gold medal at the 13th International Invention Fair in the Middle East (IIFME). Additionally, he has been listed among the Top 2% of scientists in the world by Stanford University.

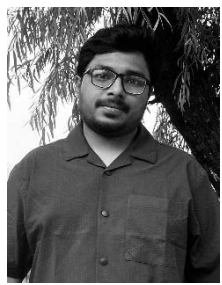
**Mir Hamidul Hussain** is working as an adjunct lecturer in the Department of Computer Science and Engineering at BRAC University. He is also a Research Assistant in the Machine Learning Group at Qatar University. He received his bachelor degree in Electrical and Electronic Engineering from University of Dhaka. His research interests are Machine learning, Computer Vision, Biomedical Image Processing and IoT.

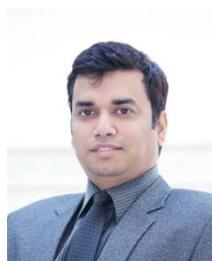
**Dr. Saad Kashem** is a Senior Lecturer and D. Program Leader in Computing Science at the AFG College with the University of Aberdeen. He has 14 years of experience as an Instructor, Course Coordinator, and Researcher of Robotics, Computer Science and Engineering courses in Australia, Malaysia, and Qatar. He received his B.Sc. in Electrical and Electronic Engineering from East West University and Ph.D. in Robotics and Mechatronics from Swinburne University of Technology, Australia. At present, he is the Adjunct Professor at Vel Tech University and Visiting Associate Professor at Presidency University. Prior to that he worked at Qatar Armed Forces – Academic Officers Program, Qatar Foundation, Qatar University, Texas A&M University, and Swinburne University of Technology. Dr. Saad also worked in the industry for two years as an Executive Electrical Engineer and Control System Engineer in Brisbane Australia. Dr. Saad has already published 1 patent, 1 book, 5 book chapters, 47 peer-reviewed journals (Web of Science indexed:12, Scopus indexed:19), and 22 conference articles (Total= 73). https://scholar.google.com/citations?user=_Sxh6SUAAAAJ&hl=en. He is a professional member of Institution of Engineering and Technology, UK (IET), Institute of Electrical and Electronics Engineers (IEEE), IEEE Robotics and Automation Society, and International Association of Engineers (IAENG). He is an editorial board member and reviewer of many international reputed Journals such as Journal of Electrical and Electronic Engineering (USA), IEEE Transactions on Control Systems Technology (USA), Vehicle System Dynamics, Taylor and Francis Ltd, (UK) etc. His research interests include Artificial Intelligence, Robotics, Machine Learning: Bio Medical Engineering, and Renewable Energy.

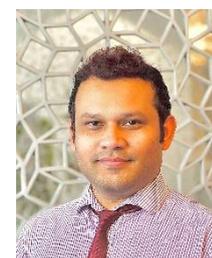
**Molla E Majid** holds a PhD in Business Intelligence from Curtin University, Australia, and dual master's degrees, one in Computing from the University of Wollongong, Australia, and another in Education from Western Sydney University, Australia. His academic journey began with a Bachelor of Engineering degree from Bangladesh University of Engineering and Technology (BUET). Dr. Majid currently serves as a faculty member at the Academic Bridge Program in Qatar Foundation, Doha. His passion for education and research has also led him to work with institutions like the Department of Education Western Australia, North South University in Dhaka, and Independent University in Dhaka, among others.

Dr. Majid's research interests include Information Systems, Business Intelligence, ICT Sustainability, Artificial Intelligence and Data Science.

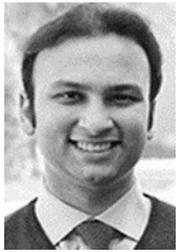 **Amith Khandakar** received a B.Sc. degree in electronics and telecommunication engineering from North South University, Bangladesh; a master's degree in computing (networking concentration) from Qatar University, in 2014 and a Ph.D. in Biomedical Engineering from UKM, Malaysia in 2023. He graduated as the Valedictorian (President Gold Medal Recipient) of North South University. He currently has around 100 journal publications, 10 book chapters, and 3 registered patents under his name. He is also a certified project management professional and a Cisco Certified Network Administrator. His research area involves sensors and instrumentation, electronics, engineering education, biomedical engineering, and machine learning applications. He has been listed among the Top 2% of scientists in the World List-2023, published by Stanford University.